\documentclass[letterpaper, 10pt, conference]{ieeeconf}
\usepackage{graphicx}
\usepackage{amsmath,amssymb}
\usepackage{algorithm}
\usepackage{algorithmic}
\usepackage{booktabs}
\usepackage{hyperref}
\usepackage{subcaption}
\usepackage[backend=biber, style=ieee, style=numeric-comp, sorting=none, labelnumber]{biblatex}
\IEEEoverridecommandlockouts

\title{
    Bayesian Optimization for Automatic Tuning of\\Torque-Level Nonlinear Model Predictive Control
}

\author{
    \authorblockN{
        Gabriele Fadini\authorrefmark{1}$^{\star}$,
        Deepak Ingole\authorrefmark{1},
        Tong Duy Son\authorrefmark{2},
        Alisa Rupenyan\authorrefmark{1}}
    \authorblockA{
        \authorrefmark{1} ZHAW Centre for Artificial Intelligence, Zürich University of Applied Sciences, Winterthur, Switzerland\\
        \authorrefmark{2} Siemens Digital Industries Software, Leuven, Belgium
    }
    \thanks{
        $^\star$ Corresponding Author
    }
    \thanks{
        \noindent This work was supported as a part of NCCR Automation, a National Centre of Competence in Research, funded by the Swiss National Science Foundation  (grant number 51NF40\_225155).
    }
    \thanks{
        {\href{mailto:fadi@zhaw.ch}{\texttt{\{fadi,inge,rupn\}@zhaw.ch}}}, {\href{mailto:son.tong@siemens.com}{\texttt{son.tong@siemens.com}}}
    }
}

\newcommand{\smallminus}{\scalebox{0.5}[1.0]{$-$}}

\begin{document}

\maketitle

\begin{abstract}
This paper presents an auto-tuning framework for torque-based Nonlinear Model Predictive Control (nMPC),
where the MPC serves as a real-time controller for optimal
joint torque commands.
The MPC parameters, including cost function weights and
low-level controller gains,  are optimized using high-dimensional
Bayesian Optimization (BO) techniques, specifically Sparse Axis-Aligned Subspace (SAASBO)
with a digital twin (DT) to achieve
precise end-effector trajectory real-time tracking on an UR10e robot arm. The simulation model allows efficient
exploration of the high-dimensional parameter space, and it ensures safe transfer to hardware.
Our simulation results demonstrate significant improvements in
tracking performance (+41.9\%) and reduction in solve times (-2.5\%) compared to manually-tuned parameters.
Moreover, experimental validation on the real robot follows the trend (with a +25.8\% improvement), emphasizing the importance of
digital twin-enabled automated parameter optimization for robotic operations.
\end{abstract}


\section{Introduction}
\label{sec:intro}
Torque-based Model Predictive Control (MPC) has emerged as a powerful
framework for robot control, enabling the direct selection of joint
torques while planning optimal control sequences over a receding
horizon~\cite{rawlings_model_2017}. Unlike kinematic controllers that
rely on cascaded control loops, torque MPC computes optimal torque
commands respecting actuator limits and dynamic constraints
\cite{budhiraja_differential_2018, tassa_control-limited_2014,
mastalli_crocoddyl_2020}.

\begin{figure}[tbp]
    \centering
    \includegraphics[width=0.5\linewidth]{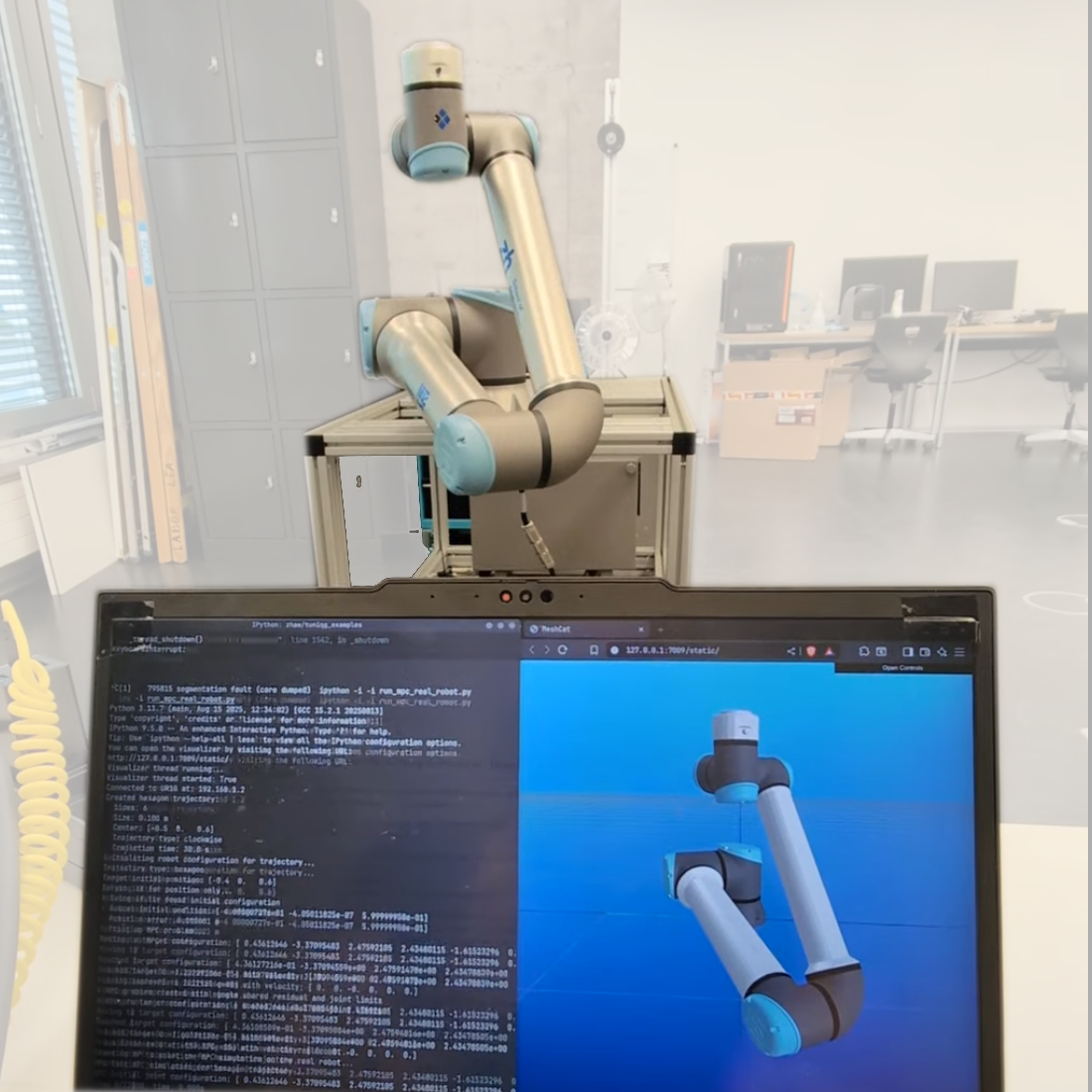}
    \caption{UR10e robot executing torque-based MPC
    leveraging a digital twin for real-time optimization.}
    \label{fig:ur10e_mpc}
\end{figure}

The practical success of torque MPC depends critically on tuning
its many parameters, such as the weights in the optimization problem, solver tolerances,
and the low-level controller feedback gains.
Each combination creates different trade-offs between tracking accuracy,
computational efficiency, and robustness. While MPC provides a solid foundation for optimal control, realizing its full potential
requires systematic refinement of these parameters \cite{rupenyan_performance-based_2021}.
Manual tuning of this high-dimensional space is tedious, often suboptimal,
and highly task-dependent.
Unlike kinematic control, torque-level MPC enables higher compliance, making
it suitable for contact-rich tasks and impedance
control~\cite{hogan_impedance_1985}, but this formulation
can potentially increase complexity
and the chance of system modeling error.
Hence, torque-level MPC requires accurate dynamics
modeling to ensure successful sim-to-real transfer~\cite{perrot_step_2023}.
We address this challenge by leveraging a system's digital
twin for safe MPC parameter exploration (Fig.~\ref{fig:ur10e_mpc}), combined
with automated tuning methods~\cite{edwards_automatic_2021,
puigjaner_performance-driven_2025, guzman_bayesian_2022}.
Recent advances in Bayesian Optimization (BO) provide promising avenues for
automated parameter tuning~\cite{snoek_practical_2012,
rupenyan_performance-based_2021, nobar_guided_2025,
widmer_tuning_2023}. In particular, Sparse Axis-Aligned Subspace Bayesian
Optimization (SAASBO)~\cite{eriksson_high-dimensional_2021,
eriksson_scalable_2021} has shown remarkable performance in
high-dimensional problems (hundreds of parameters) by exploiting
low-dimensional structure, making it well-suited for robotic applications.

This paper's contributions are the following:
\begin{itemize}
    \item Implementation of a torque-level nMPC interfaced with an extensible
    digital twin of the UR10e robot arm.
    \item Automated parameter optimization framework using high-dimensional
    Bayesian Optimization to balance real-time execution and control
    performance.
    \item Comprehensive testing demonstrating
    improvement over baselines in simulation and the real system.
\end{itemize}

\section{Problem Formulation}

\subsection{Robot Dynamics}

Building upon prior force control methodologies \cite{jordana_force_2024, kleff_high-frequency_2021},
we augment the classical MPC formulation with an explicit model of the robot's actuation dynamics.
We consider a robot manipulator operating under torque control, where the control input
$\mathbf{u} \in \mathbb{R}^{n_u}$ consists of joint torque commands and $n_u$ is the number of joints.
The state of the robot is represented by $\mathbf{x} = [\mathbf{q}; \mathbf{v}] \in \mathbb{R}^{n_q + n_v}$, where the joint positions
$\mathbf{q} \in \mathbb{R}^{n_q}$ and velocities $\mathbf{v} \in \mathbb{R}^{n_v}$.
The robot dynamics follow the
standard manipulator equation:
\begin{equation}
    \mathbf{M}(\mathbf{q})\dot{\mathbf{v}} + \mathbf{C}(\mathbf{q},\mathbf{v})\mathbf{v} + \mathbf{g}(\mathbf{q}) = \mathbf{u},
    \label{eq:robot_dyn}
\end{equation}
where $\mathbf{M}(\mathbf{q}) \in \mathbb{R}^{n_v \times n_v}$ is the joint inertia matrix, $\mathbf{C}(\mathbf{q},\mathbf{v})$ represents Coriolis and centrifugal terms,
$\mathbf{g}(\mathbf{q})$ is the gravity vector, and $\mathbf{u}$ are the applied joint torques.

\subsection{Model Predictive Control Solver Description}
\label{sec:mpc_solver}

We formulate the torque-based MPC optimal control problem at each time step
$t \in [0,N]$ with discretization $\Delta t$ as:
\begin{equation}
\begin{aligned}
    \min_{\mathbf{u}_0,\ldots,\mathbf{u}_{N - 1}} \quad & \sum_{k=0}^{N-1} \ell_{k}(\mathbf{x}_k, \mathbf{u}_k) \Delta t + \ell_N(\mathbf{x}_N) \\
    \text{s.t.} \quad & \mathbf{x}_{k+1} = f(\mathbf{x}_k, \mathbf{u}_k), \; k=0,\ldots,N \smallminus1 \\
    & \mathbf{x}_0 = \mathbf{x}(t) ,  \\
\end{aligned}
\label{eq:ocp_prob}
\end{equation}
where $f(\mathbf{x}_k, \mathbf{u}_k)$ is an Euler integrator computing the next state enforcing the system
acceleration computed from~\eqref{eq:robot_dyn}
and $\ell_k(\mathbf{x}_k, \mathbf{u}_k)$ are the cost components to minimize. The first control $\mathbf{u}_0^\star$
is applied, then the horizon shifts forward.
In addition to the dynamics and cost in \eqref{eq:ocp_prob}, for robotic
systems, the MPC planner must generate feasible trajectories that achieve the desired motion while
respecting actuator and state limits approximated with the linear bounds:
\begin{equation}
\begin{aligned}
    \mathbf{q}_{\min} &\leq \mathbf{q}(t) \leq \mathbf{q}_{\max} \quad &&\text{Joint positions} \\
    \mathbf{v}_{\min} &\leq \mathbf{v}(t) \leq \mathbf{v}_{\max} \quad &&\text{Joint velocities}\\
    \mathbf{u}_{\min} &\leq \mathbf{u}(t) \leq \mathbf{u}_{\max} \quad &&\text{Torques},
\end{aligned}
\label{eq:constraints}
\end{equation}
where $(\cdot)_{\min}$ and $(\cdot)_{\max}$ denote the allowable limits.

We solve \eqref{eq:ocp_prob} using Differential Dynamic Programming
(DDP)~\cite{mayne_second-order_1966, mastalli_crocoddyl_2020}, which
efficiently exploits the time-sparse structure of the optimal control
problem (linear with the number of problem nodes $\mathcal{O}(N)$). Given nominal trajectories $(\bar{\mathbf{x}}_t,
\bar{\mathbf{u}}_t)_{t=0}^N$, DDP iteratively constructs a local quadratic
approximation and improves the control sequence via backward-forward
passes. The backward pass computes second-order value function
approximations, yielding feedforward gains $\mathbf{k}_t$ and feedback gains
$\mathbf{K}_t$. The forward pass then applies the improved control guess in simulation,
locally improving the trajectories of states and control inputs for the next backward pass.
At convergence, a locally optimal solution is achieved.
Among other variants of DDP, Crocoddyl's Feasibility-driven DDP (FDDP)
~\cite{mastalli_crocoddyl_2020}, enables real-time performance
through the relaxation of the constraints \eqref{eq:constraints} as penalties,
Hessian Newton approximation, and warm-starting capabilities.

\subsection{Motivation for Torque-Level Model Predictive Control}

Traditional kinematic-based MPC operates at the position-velocity level, abstracting
dynamics for faster computation, commanding joint positions and velocities,
but limiting compliant manipulation and force
control~\cite{hogan_impedance_1985, wahrburg_mpc-based_2016}. Torque-based
MPC directly optimizes joint torques as decision variables, enabling contact-rich
tasks~\cite{mastalli_crocoddyl_2020}, compliant
behavior~\cite{kleff_high-frequency_2021}, and direct actuator
optimization~\cite{tassa_control-limited_2014}. However, incorporating
full dynamics (and joint torques as decision variables) increases the computational complexity of the DDP problem from $\mathcal{O}((2n_q)^3 N)$ to
$\mathcal{O}((3n_q)^3 N)$, requiring careful tuning of cost weights, solver
tolerances, and feedback gains. This challenge lacks established heuristics,
making systematic parameter optimization essential.

\section{Bayesian Optimization Framework}

\subsection{Automatic Tuning Overview}

\begin{figure}[tbp]
    \centering
    \includegraphics[width=0.75\linewidth]{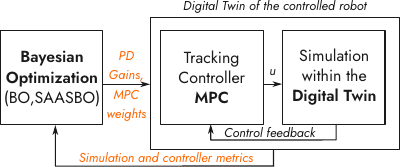}
    \caption{Overview of the Bayesian Optimization, a digital twin is used to evaluate the MPC.}
    \label{fig:loop_bayesian}
\end{figure}

Algorithm~\ref{alg:saasbo} presents our MPC hyperparameter auto-tuning framework.
Our algorithm proceeds in two phases. First, an initial dataset is
generated through Latin Hypercube sampling to uniformly explore the
parameter space. For each sampled parameter vector $\boldsymbol{\theta}_i$,
the MPC is executed in the digital twin, collecting the resulting
trajectory data $(\mathbf{x}, \mathbf{u})$ and computing the objective
function $J(\boldsymbol{\theta}_i)$.
This initial exploration phase produces $n_0$ observation pairs $\mathcal{D} = \{(\boldsymbol{\theta}_i, y_i)\}_{i=1}^{n_0}\}$.
Given such observations, a Gaussian process models the relationship between
$\boldsymbol{\theta}$ and $y$:
\begin{equation}
    y \sim \mathcal{GP}(m(\boldsymbol{\theta}), k(\boldsymbol{\theta}, \boldsymbol{\theta}')).
\end{equation}
A Matérn-5/2 kernel is used, assuming $J$ is twice differentiable:
\begin{equation}
    k(\boldsymbol{\theta}, \boldsymbol{\theta}') = \sigma^2 \left(1 + \sqrt{5}r + \frac{5}{3}r^2\right) \exp(-\sqrt{5}r),
\end{equation}
where $r = \sqrt{\sum_{d=1}^D (\boldsymbol{\theta}_d - \boldsymbol{\theta}'_d)^2/\ell_d^2}$
and $\ell_d$ are length scales learned via Markov Chain Monte Carlo (MCMC).

\begin{algorithm}[tbp]
\caption{MPC Parameters Optimization}
\begin{algorithmic}[1]
\STATE Initialize hyperparameter bounds $[\boldsymbol{\theta}_{\min}, \boldsymbol{\theta}_{\max}]$
\STATE Generate $n_0$ Latin Hypercube samples
\FOR{$i = 1$ to $n_0$}
    \STATE Run MPC with parameter $\boldsymbol{\theta}_i$ with digital twin
    \STATE Collect trajectory data $\mathbf{x}, \mathbf{u}$
    \STATE Evaluate metric $y_i = J(\boldsymbol{\theta}_i)$
\ENDFOR
\STATE $\mathcal{D} \gets \{(\boldsymbol{\theta}_i, y_i)\}_{i=1}^{n_0}$
\STATE Initialize GP model based on $\mathcal{D}$
\FOR{$t = n_0+1$ to $n_{\max}$}
    \STATE $\boldsymbol{\theta}_t \gets \arg\max_{\boldsymbol{\theta}} \text{EI}(\boldsymbol{\theta})$
    \STATE Evaluate metric $y_t = J(\boldsymbol{\theta}_t)$
    \STATE $\mathcal{D} \gets \mathcal{D} \cup \{(\boldsymbol{\theta}_t, y_t)\}$
    \STATE Refine GP model with $\mathcal{D}$
    \IF{no improvement for $p$ iterations}
        \STATE Break (early stopping)
    \ENDIF
\ENDFOR
\RETURN Optimal parameters: $\boldsymbol{\theta}^* = \arg\min_{\boldsymbol{\theta} \in \mathcal{D}} J(\boldsymbol{\theta})$
\end{algorithmic}
\label{alg:saasbo}
\end{algorithm}

While in theory any Bayesian Optimization (BO) technique could be used in Algorithm~\ref{alg:saasbo},
they would struggle with the curse of dimensionality.
As a solution, we exploit SAASBO, which is particularly well-suited for high-dimensional tuning.
SAASBO identifies a low-dimensional structure within the parameter space.
Unlike standard BO, which treats all dimensions equally, SAASBO
employs hierarchical sparsity-inducing priors (half-Cauchy distributions)
on the inverse lengthscales of the GP kernel~\cite{paananen_variable_2019}.
This encourages most parameters to have minimal influence while identifying
a few critical dimensions, effectively reducing the search to a low-dimensional
subspace.
The method uses Hamiltonian Monte Carlo with No-U-Turn Sampler (NUTS) to identify these critical
dimensions, offering three key advantages: (i) sample efficiency through
selective modeling, (ii) preserved input geometry for better constraint
handling, and (iii) robust scaling to hundreds of dimensions, making it ideal
for the expensive MPC evaluations (each in the order of 10s).

At every iteration, the sparse $\mathcal{GP}$ model identifies
which regions of the reduced parameter space to explore.
The Expected Improvement (EI)
acquisition function is used:
\begin{equation}
    \text{EI}(\boldsymbol{\theta}) = \mathbb{E}[\max(f^* - f(\boldsymbol{\theta}), 0)],
\end{equation}
where $f^*$ is the current best observed value.
The next evaluation point chosen by maximizing the
EI:
\begin{equation}
    \boldsymbol{\theta}_{n+1} = \arg\max_{\boldsymbol{\theta}} \text{EI}(\boldsymbol{\theta}).
\end{equation}
At every iteration, a new observation is added to the dataset $\mathcal{D}$,
refining the $\mathcal{GP}$.
The optimization terminates when no improvement is observed over $p$ consecutive iterations,
or when a maximum iteration budget $n_{\max}$ is reached.
The best achieved parameter configuration $\boldsymbol{\theta}^\star$ during the optimization is finally selected.

\section{Implementation Details}

\subsection{Digital Twin for Parameter Optimization}

Each iteration of Algorithm~\ref{alg:saasbo} requires evaluating the MPC planner with different hyperparameters $\boldsymbol{\theta}$,
as shown in Fig.~\ref{fig:loop_bayesian}.
This can be sped up \textit{in silico}, making automated
optimization feasible.
A digital twin of the UR10e robot, built using Pinocchio \cite{carpentier_pinocchio_2019},
with a parameterized URDF model, enables safe exploration of the MPC hyperparameter
space.
The digital twin also enables safe testing, avoiding failure modes
(joint limits, actuator saturation) that would be dangerous on the physical system.
The fidelity of the digital twin to the real robot is validated, after calibration of the
end-effector. We verify that the kinematic position estimation of the end-effector matches, within numerical precision $\approx 10^{\smallminus 16}$, the one estimated by the robot internal functions.
Also, the values of the computed joint mass inertia matrix $\mathbf{M}(\mathbf{q})$ are compared to those obtained
from the real robot for 100 random joint configurations, showing an average error of 2$\cdot10^{\smallminus 5}$ (Frobenius norm).
Our digital twin hence captures well kinematic and dynamic properties but, due to
the lack of external sensors, we cannot encompass
friction, backlash, or communication delays.

\subsection{Low-Level Joint Controller}

\begin{figure}[tbp]
    \centering
    \includegraphics[width=0.9\linewidth]{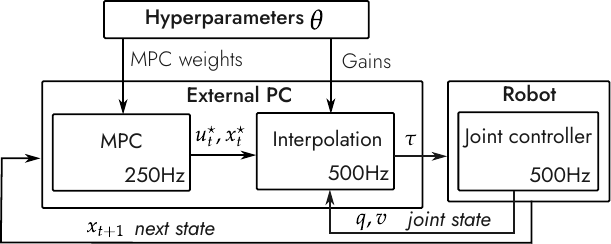}
    \caption{Low-level torque control interface for UR10e robot using RTDE library.}
    \label{fig:control_loop}
\end{figure}

To bridge the real-time gap between the MPC frequency (125-250~Hz due to
computational constraints) and the required 500~Hz robot control frequency, we employ joint-space
and task-space feedback. At each MPC update, the solver provides an optimal torque trajectory
$\{\mathbf{u}_0^\star, \mathbf{u}_1^\star, \ldots, \mathbf{u}_{N-1}^\star\}$ along with the associated optimal state trajectory $\{\mathbf{x}_0^\star, \mathbf{x}_1^\star, \ldots, \mathbf{x}_{N-1}^\star\}$.
Between MPC updates, torque commands are interpolated at higher frequency, and corrected with joint and task feedback components:
\begin{equation}
    \begin{aligned}
    \mathbf{\tilde{u}}(t) \; = \; \mathbf{u}_k^\star &+ \mathbf{K}_d (\mathbf{v}_{k+1}^\star - \mathbf{v}(t))
                                                + \mathbf{K}_p (\mathbf{q}_{k+1}^\star - \mathbf{q}(t))\\
                                               &+ \mathbf{K}_{p,c} \mathbf{J}_k^{\top}(\mathbf{p}(\mathbf{q}^\star_{k+1}) - \mathbf{p}(\mathbf{q}(t)))\\
                                               &+ \mathbf{K}_{d,c} \mathbf{J}_k^{\top}(\mathbf{J}_{k+1}(\mathbf{v}^\star_{k+1}) -\mathbf{J}_{k}(\mathbf{v}(t))),
    \end{aligned}
    \label{eq:low_level_control}
\end{equation}
where, at every time step $k$, $\mathbf{u}_k^\star$ and $\mathbf{v}_{k+1}^\star$ are the optimal torque and velocity from the MPC solution, and $\mathbf{K}_p,\mathbf{K}_d \in \mathbb{R}$ are respectively joint position and velocity feedback gain.
By computing $\mathbf{p}, \mathbf{v}, \mathbf{J}_{\boldsymbol{\cdot}}$, respectively, the end-effector's position velocities, and the joint Jacobian, we can also add a task-dependent feedback
where $\mathbf{K}_{p,c},\mathbf{K}_{d,c} \in \mathbb{R}^3$ are the cartesian position and velocity gains.

This hybrid feedback-feedforward controller structure is
critical for real-world deployment to maintain more reliable tracking performance.
This controller formulation strikes a balance between expressiveness, interpretability,
and a reduced number of governing parameters, which is desirable for automated tuning.

The joint torque commands $\boldsymbol{\tau}$ computed after interpolation are executed on the UR10e robot using the
direct torque control API introduced in the UR10e firmware and the \texttt{RTDE 1.6.2} library \cite{lindvig_ur_rtde_2025}.
This interface
provides low-level access
to joint-level torque control \cite{nvidia_tech_report_2023}.
The UR10e torque controller additionally
compensates for gravity forces $\mathbf{g}(\mathbf{q})$ acting on the robot as in \eqref{eq:robot_dyn}.
The command sent to the robot, to enforce safety, is additionally saturated and compensated:
\begin{equation}
    \boldsymbol{\tau} = \text{sat}(\mathbf{\tilde{u}}, \mathbf{u}_{\min}, \mathbf{u}_{\max}) - \mathbf{g}(\mathbf{q}),
\end{equation}
where $\text{sat}(\cdot, \mathbf{u}_{\min}, \mathbf{u}_{\max})$ denotes element-wise saturation to ensure
commanded torques remain within the actuator limits.

\subsection{MPC Formulation}

\begin{figure}[tbp]
    \centering
    \begin{subfigure}{0.32\linewidth}
        \centering
        \includegraphics[width=\linewidth]{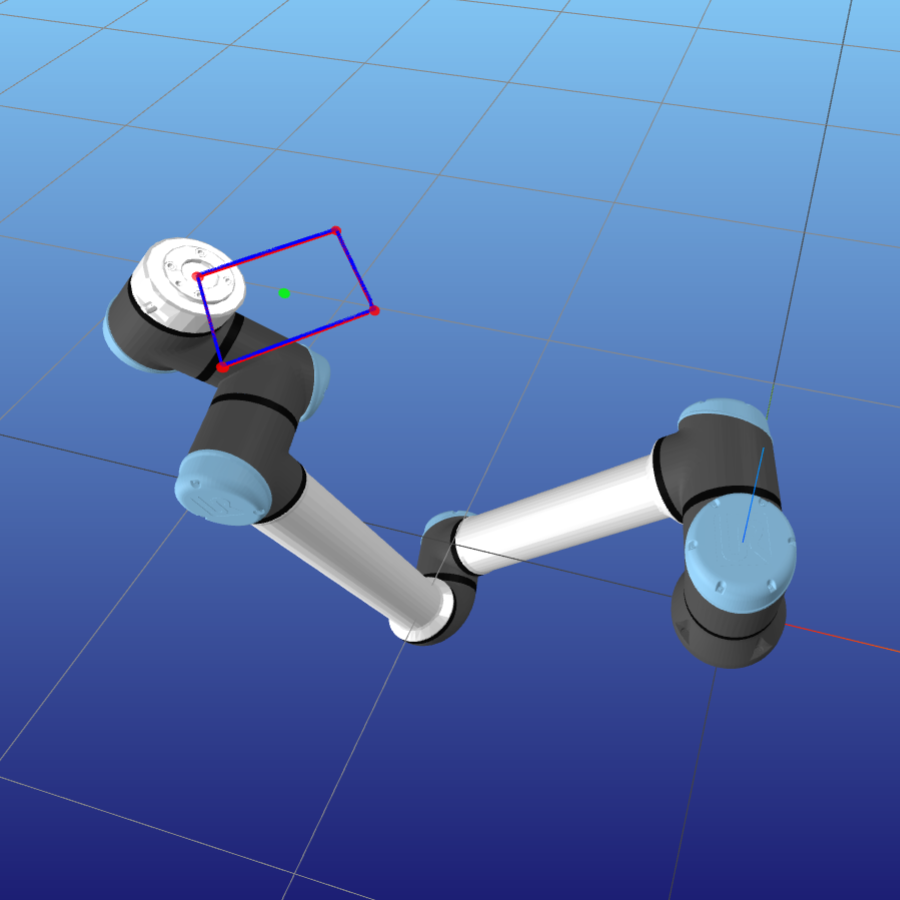}
        \caption{Square}
    \end{subfigure}
    \begin{subfigure}{0.32\linewidth}
        \centering
        \includegraphics[width=\linewidth]{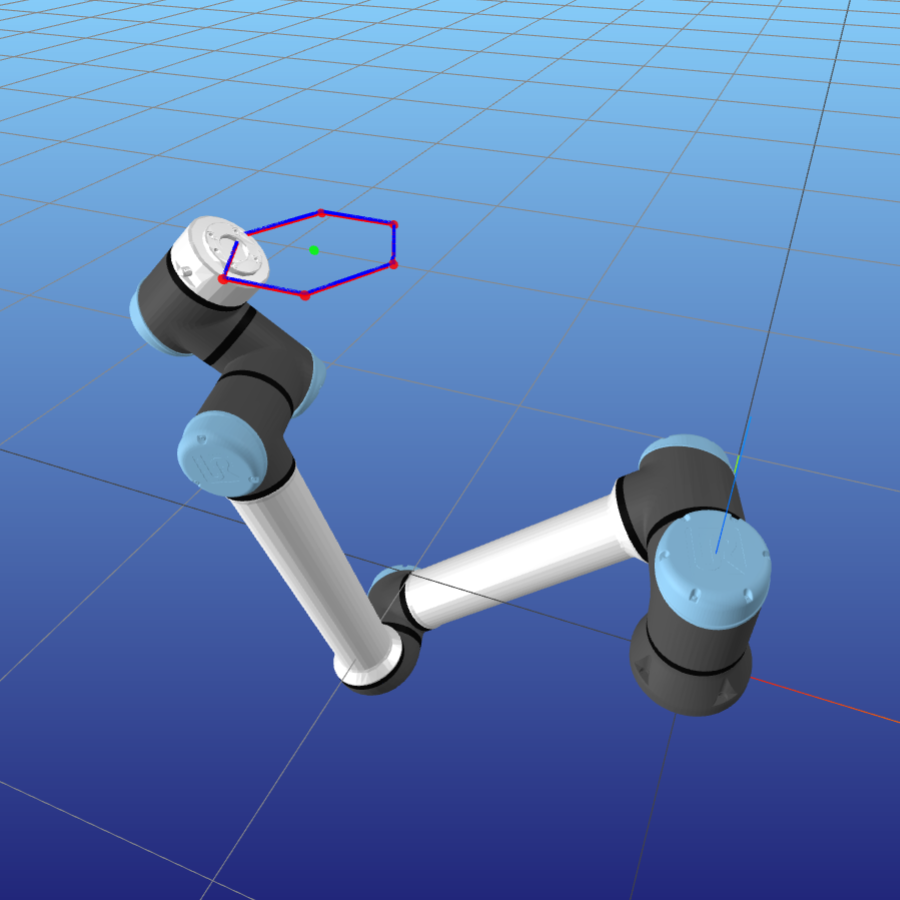}
        \caption{Hexagon}
    \end{subfigure}
    \begin{subfigure}{0.32\linewidth}
        \centering
        \includegraphics[width=\linewidth]{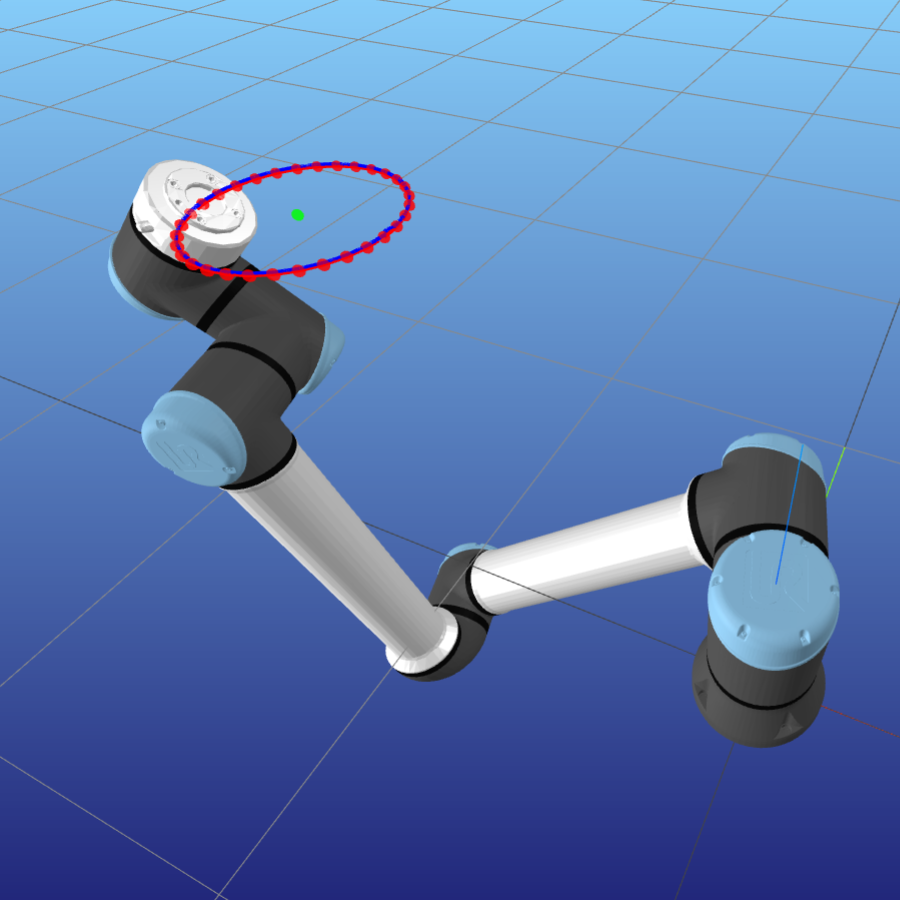}
        \caption{Circle}
    \end{subfigure}
    \caption{Tracking MPC trajectories in task-space.}
    \label{fig:trajectories}
\end{figure}

\noindent \textit{Tracking Task:} The control objective is to track a reference
end-effector trajectory $\mathbf{p}_{\text{des}}(t)$ while minimizing tracking
error, control effort, and joint limits. We generate trajectories
(square, hexagon, circle) with parameterized size and center as in
Fig.~\ref{fig:trajectories}, with reference point $\mathbf{p}_\text{des}(t)$ and end-effector
orientation $\mathbf{R}_\text{des}(t)$ being
linearly interpolated between vertices based on a task-progress variable $\phi \in [0,1]$.

Omitting the time index $k$, at every node, the running cost $\ell(\mathbf{x},\mathbf{u})$ takes the form:
\begin{equation}
\begin{aligned}
    \ell(\mathbf{x},\mathbf{u}) = \; & w_{\text{pos}} \|\mathbf{p}(\mathbf{q}) - \mathbf{p}_{\text{des}}\|^2 +
    w_{\text{rot}} \|\mathbf{R}(\mathbf{q}) - \mathbf{R}_{\text{des}}\|^2_F + \\
    & w_{\tau} \|\mathbf{u}\|^2 + w_{\text{v}} \|\mathbf{v}\|^2 + \\
    & w_{\text{lim}, \tau} \Gamma(\mathbf{u}) + w_{\text{lim}, x} \Gamma(\mathbf{x}),
\end{aligned}
\label{eq:mpc_formulation}
\end{equation}
where $\mathbf{p}$ and $\mathbf{R}$ represent the end-effector position and orientation,
$\Gamma(\mathbf{u})$ and $\Gamma(\mathbf{x})$ are quadratic barrier functions enforcing control and actuator limits
(velocity and position),
and $w_{\boldsymbol{\cdot}}$ are tunable MPC weights
that determine the trade-off between tracking accuracy, control smoothness, and constraint satisfaction.
The terminal cost adds ${w_{\text{pos,N}} \|\mathbf{p}(\mathbf{q}_N) - \mathbf{p}_{\text{des,N}}\|^2}$ at the horizon end $N$,
with $w_{\text{pos,N}}$ coupled to $w_{\text{pos}}$ to maintain consistency between running and terminal objectives.
Table~\ref{tab:mpc_weights} summarizes the nominal cost weights used in the MPC formulation.

All cost weights except the barrier-dependent ones are tunable
hyperparameters for optimization.
In the decision variables $\boldsymbol{\theta} \in \mathbb{R}^{12}$ we also include the joint and task gains:
\begin{equation}
    \boldsymbol{\theta} = [w_{\text{pos}}, w_{\text{rot}}, w_\tau, w_{\text{v}}, \mathbf{K}_p, \mathbf{K}_d, \mathbf{K}_{p,c}, \mathbf{K}_{d,c}],
\end{equation}
where the different terms have been defined in \eqref{eq:low_level_control} and \eqref{eq:mpc_formulation}.
The objective function balances optimality with real-time performance with
a convex combination depending on a tunable parameter $\alpha \in [0, 1]$:
\begin{equation}
    J(\boldsymbol{\theta}) = \; \alpha \cdot \mathcal{L} \; +
                                (1 - \alpha) \cdot {t}_{\text{solve}} ,
\end{equation}
where $\mathcal{L}$ is the accumulated cost over the trajectory as in~\eqref{eq:ocp_prob},
and $t_{\text{solve}}$ is the average DDP solve time.
In our experiments, we set $\alpha = 0.8$ to prioritize tracking performance.
The optimization seeks parameters that enable the MPC to generate torques,
achieving minimal tracking error while ensuring the DDP solver
converges within real-time constraints.

\begin{table}[tbp]
\centering
\caption{Model Predictive Control Cost Weights}
\label{tab:mpc_weights}
\begin{tabular}{lccc}
\toprule
\textbf{Cost Components} & \textbf{Default} & \multicolumn{2}{c}{\textbf{Optimized}}\\
\midrule
\textit{Running Costs Weights} & & Vanilla BO & SAASBO\\
\; Position tracking, $w_{\text{pos}}$ & $10^5$ & $7.2 \times 10^4$ & $4.1 \times 10^4$\\
\; Rotation tracking, $w_{\text{rot}}$ & $10^{\smallminus 4}$ & $5.8 \times 10^{\smallminus 5}$ & $2.3 \times 10^{\smallminus 5}$\\
\; Control reg., $w_{{\tau}}$ & $10^{\smallminus 2}$ & $6.5 \times 10^{\smallminus 3}$ & $3.7 \times 10^{\smallminus 3}$\\
\; Joint velocities, $w_{\text{v}}$ & $10^{\smallminus 3}$ & $8.1 \times 10^{\smallminus 4}$ & $7.9 \times 10^{\smallminus 4}$\\
\textit{Fixed Running Cost Weights} & & & \\
\; Control limits, $w_{\text{lim}, \tau}$ & $10^{1}$ & $10^{1}$ & $10^{1}$\\
\; Joint limits, $w_{\text{lim}, x}$ & $10^{1}$ & $10^{1}$ & $10^{1}$\\
\midrule
\textit{Terminal Cost Weight} & & & \\
\; Position tracking, $w_{\text{pos,N}}$ & $w_{\text{pos}}$ & $w_{\text{pos}}$ & $w_{\text{pos}}$\\
\bottomrule
\end{tabular}
\end{table}


\subsubsection{Solver Configuration and Real-Time Control}
The reactive controller MPC employs FDDP from Crocoddyl~\cite{mastalli_crocoddyl_2020} with a horizon of
$N=20$ steps (50ms lookahead), maximum 10 iterations per solver call, targeting $\leq$2ms solve time
for 500~Hz control. At every cycle the solver is warm-started from previous solutions.

\subsection{High-dimensional Bayesian Optimization Configuration}
\label{sec:bo_config}

For the SAASBO optimization, we configure the following settings and
implement the algorithm using the BoTorch library~\cite{balandat_botorch_2020}
and AX platform~\cite{olson_ax_2025}. The optimization begins with $n_0=100$
initial samples generated via Latin Hypercube sampling to ensure uniform
coverage of the 12-dimensional parameter space, followed by 200 Bayesian Optimization
iterations. Each MPC evaluation requires $\approx$~15$\smallminus$20~s in the digital twin,
making the total parameter tuning feasible in 1h06min. The early stopping threshold
$p=10^2$ ensures thorough exploration improving the $\mathcal{GP}$ model, though convergence
typically occurs earlier. The Gaussian process hyperparameters are inferred using MCMC
with 1024 warmup steps and 1024 posterior samples (thinned by 16) for
robust uncertainty quantification. The kernel uses smoothness $\nu = 2.5$
with output scale $\sigma^2$ and lengthscales $\ell_d$ learned via MCMC,
while a fixed noise variance $10^{\smallminus 6}$ reflects the
near-deterministic simulation.

\section{Experimental Results}

\begin{figure}[tbp]
    \centering
    \includegraphics[width=0.8\linewidth]{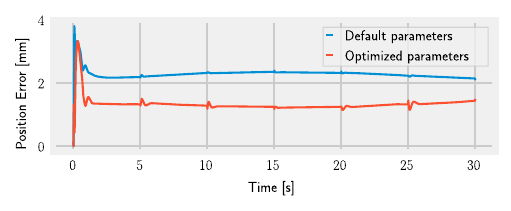}
    \caption{Tracking error trend, the baseline is compared with optimized parameters.}
    \label{fig:simulation_improvement}
\end{figure}

\subsection{Automatic Tuning Results in Simulation}
\label{subsec:simulation_results}

We run the optimization loop (Algorithm~\ref{alg:saasbo}), evaluating the MPC planner on the digital twin for each sampled
parameter set in a task in which the UR10e robot tracks a hexagonal
trajectory (Fig.~\ref{fig:trajectories}) of 10cm side length in a total
time of 30~s. The optimization
 yields a set of parameters that
significantly improves tracking performance, while it maintains reliable real-time control performance, compared to both the manually adjusted default
 baseline and a vanilla BO approach using the same seed,
configuration, and initial population as described in Section~\ref{sec:bo_config}.
\begin{table}[tbp]
\centering
\caption{Trajectory End-effector Tracking in Simulation}
\label{tab:performance_summary}
\begin{tabular}{lcc}
\toprule
\textbf{Method} & \textbf{Avg Error (mm)} & \textbf{Max Error (mm)} \\
\midrule
Baseline & 2.29 $\pm$ 0.15 & 3.81 \\
Vanilla BO & 1.61 $\pm$ 0.18 & 3.57 \\
SAASBO & 1.33 $\pm$ 0.22 & 3.33 \\
\midrule
Improvement (Vanilla BO) & +29.7\% & +6.3\%\\
Improvement (SAASBO) & +41.9\% & +12.5\%\\
\bottomrule
\end{tabular}
\end{table}

\begin{table}[tbp]
\centering
\caption{Optimized Feedback Gains}
\label{tab:feedback_gains}
\begin{tabular}{lcccc}
\toprule
\textbf{Control Level} & \textbf{Gain} & \textbf{Default} & \textbf{Vanilla BO} & \textbf{SAASBO} \\
\midrule
Joint-space & $\mathbf{K}_p$ & $1$ & $12.3$ & $28.7$ \\
Joint-space & $\mathbf{K}_d$ & $1$ & $0.45$ & $0.18$ \\
\midrule
Task-space & $\mathbf{K}_{p,c}$ & $[1, 1, 1]$ & $[3.2, 2.8, 15.4]$ & $[7.8, 6.5, 89.2]$ \\
Task-space & $\mathbf{K}_{d,c}$ & $[1, 1, 1]$ & $[1.5, 1.3, 8.7]$ & $[2.1, 1.8, 10.3]$ \\
\bottomrule
\end{tabular}
\end{table}

\subsubsection{Weights Analysis}
Fig.~\ref{fig:simulation_improvement} illustrates the reduction in tracking error over the optimization process.
Table~\ref{tab:performance_summary} presents the tracking performance comparison.
Tab.~\ref{tab:mpc_weights} compares the cost weights across methods, with the
SAASBO column showing the most effective configuration.
The position tracking weight $w_{\text{pos}}$ is reduced by an order of magnitude,
indicating that overly aggressive tracking can lead to instability or excessive control effort.
The control regularization weight $w_{\tau}$ is also decreased, allowing for more responsive torque commands.
The rotation tracking weight $w_{\text{rot}}$ is lowered, suggesting that precise orientation tracking is less critical for the task.
Additionally, with optimized weights, the MPC solver achieves a modest $2.55\%$ reduction in the average
number of DDP iterations (on average 1.009 iterations in 2 ms),
enabling consistent real-time exectution.
\subsubsection{Gain Analysis}
The trade-off between feedforward MPC torque and feedback terms
is non-trivial, and understanding optimal gain magnitudes while avoiding instability
is aided by the use of the digital twin to predict performance.
The optimized hybrid controller increases the feedback contribution for tracking.
Table~\ref{tab:feedback_gains} summarizes the gains used for the controller.
Joint stiffness increases significantly
while damping decreases, suggesting the system benefits from aggressive proportional
tracking with minimal derivative action. 
Task-space gains show substantial increases in the z-axis (vertical):
compared to xy-plane values, reflecting the need for higher vertical stiffness for a precise
lateral motion execution.
\subsubsection{Comparison with Vanilla Bayesian Optimization}
\begin{figure}[tbp]
    \centering
    \includegraphics[width=0.6\linewidth]{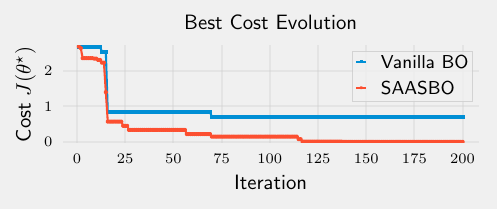}
    \caption{Comparison of the optimization progress (best cost from parameters $\boldsymbol{\theta}^\star$) between SAASBO and vanilla BO.}
    \label{fig:bayesian_optimization_improvement}
\end{figure}

Starting from the same priors and an initial cost of 2.677, SAASBO achieves a final cost of 0.23 after 200 iterations (91.3\%
improvement), while vanilla BO reaches 0.703 (29.7\% improvement). This substantial gap arises because
vanilla BO treats all 12 parameters equally, spreading its modeling capacity across the full space.
In contrast, SAASBO's sparsity-inducing priors effectively identify that only 3-5 parameters dominate
performance (primarily $w_{\text{pos}}$, $K_p$, and z-axis task gains), enabling focused exploration
of the relevant subspace and reducing the effect of the curse of dimensionality.
In Fig.~\ref{fig:bayesian_optimization_improvement}, the trend of the current best
cost value over iterations is shown for both methods, highlighting how
SAASBO consistently improves throughout the optimization, while vanilla BO
returns a worse solution within the computation constraints.
\subsection{Experimental testing on the real robot}
\label{subsec:robot_results}
\begin{figure}[tbp]
    \centering
    \begin{subfigure}[c]{0.59\linewidth}
        \centering
        \includegraphics[width=\linewidth]{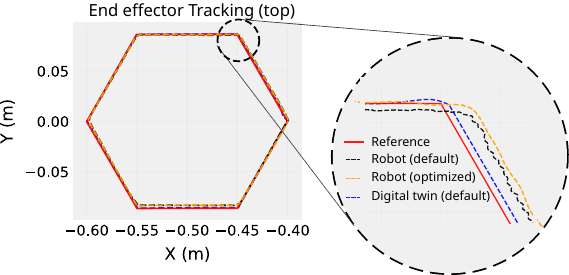}
        \caption{Error in task-space}
        \label{fig:tracking_comparison}
    \end{subfigure}
    \begin{subfigure}[c]{0.39\linewidth}
        \centering
        \includegraphics[width=\linewidth]{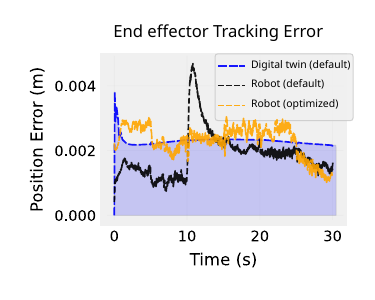}
        \caption{Error over time}
        \label{fig:tracking_error}
    \end{subfigure}
\caption{End-effector cartesian tracking error in simulation and on the robot for the hexagon trajectory.}
\label{fig:optimization_sim_to_real}
\end{figure}

Hardware validation of parameters found in Sec.~\ref{subsec:simulation_results} was conducted on the UR10e robot tracking the same
hexagonal trajectory as in Sec.~\ref{subsec:simulation_results}.
Fig.~\ref{fig:optimization_sim_to_real} shows
the tracking performance across three configurations: default parameters in
the digital twin (blue), default parameters on the real robot
(black), and optimized parameters on the real robot (yellow).
Fig.~\ref{fig:tracking_comparison} shows the Cartesian tracking error in
task space. The optimized parameters produce trajectories (yellow)
that remain generally closer to the reference path (red) and exhibit fewer
high-frequency oscillations compared to the ones from default parameters (black).
The results show an increase in tracking error for both
default and optimized parameters when transferring from simulation to hardware, likely due to unmodelled
effects such as friction, communication delays, and sensor noise.

The position tracking error in time, shown in Fig.~\ref{fig:tracking_error}, shows that the parameters
optimized with SAASBO achieve a lower average
tracking error (yellow) and significantly reduced error peaks compared to the default
ones (black). Quantitatively, the average tracking error on
hardware is 2.3~mm for the SAASBO optimized parameters versus the 3.1~mm achieved with the default parameters
(25.8\% improvement). The relative improvement from optimization (25.8\% on hardware vs.
41.9\% in simulation) shows that the optimal parameters translate to the system, validating the
digital twin as an effective optimization surrogate for real experimentation despite the sim-to-real gap.

\section{Conclusion and future work}

We demonstrate that systematic hyperparameter optimization is beneficial to
tune torque-based nMPC in robotic trajectory tracking tasks.
For this, an automated tuning framework was introduced, combining high-dimensional Bayesian Optimization
(SAASBO) with a digital twin to efficiently explore the MPC parameters before hardware deployment.
Our results explore torque-level nMPC coupled with reactive feedback.
We, moreover, demonstrate the importance of sparsity-inducing priors for high-dimensional optimization in robotic applications
showing that vanilla BO stagnates in finding the best configuration.
The sim-to-real transfer in Sec.~\ref{subsec:robot_results} validates the effectiveness of our method
and the digital twin as surrogate.
Our experimental validation shows that 41.9\%
tracking improvement (compared to a baseline) in simulation still translates to a 25.8\% improvement on hardware.
Further modeling of friction, delays, and sensor noise could
further reduce the current sim-to-real gap \cite{nobar_guided_2025}.
To this end, our digital twin could either encompass a more detailed multi-domain physical simulation or
incorporate a data-driven approach.
Despite our results showing promise for parameter automatic tuning, the method
may still struggle in high-dimensional settings.
We hence plan further validation of the method
for other complex tasks, where parameter sensitivity is more pronounced, such as
contact-rich manipulation.

\renewcommand*{\bibfont}{\small}
\printbibliography

\end{document}